\documentclass[sigconf,table]{acmart}
\usepackage{multirow}
\usepackage{subcaption}
\usepackage{amsmath} 
\renewcommand{\vec}[1]{\boldsymbol{#1}} 

\usepackage{graphicx}
\usepackage{epstopdf}
\begin{document}

\title{PuFace: Defending against Facial Cloaking Attacks for Facial Recognition Models}

\author{Jing Wen}
\orcid{0003-4721-5327}
\authornote{This work was finished at the University of Hong Kong in 2022 and was also included in the author's dissertation\cite{wen2022defending}.}
\email{jwen@cs.hku.hk}
\affiliation{%
    \institution{The University of Hong Kong}
    \city{Dept. of Computer Science}
}

\begin{abstract}

The recently proposed facial cloaking attacks add invisible perturbation (cloaks) to facial images to protect users from being recognized by unauthorized facial recognition models. However, we show that the "cloaks" are not robust enough and can be removed from images.

This paper introduces PuFace, an image purification system leveraging the generalization ability of neural networks to diminish the impact of cloaks by pushing the cloaked images towards the manifold of natural (uncloaked) images before the training process of facial recognition models. Specifically, we devise a purifier that takes all the training images including both cloaked and natural images as input and generates the purified facial images close to the manifold where natural images lie. To meet the defense goal, we propose to train the purifier on particularly amplified cloaked images with a loss function that combines image loss and feature loss. Our empirical experiment shows PuFace can effectively defend against two state-of-the-art facial cloaking attacks and reduces the attack success rate from 69.84\% to 7.61\% on average without degrading the normal accuracy for various facial recognition models. Moreover, PuFace is a model-agnostic defense mechanism that can be applied to any facial recognition model without modifying the model structure.

\end{abstract}




\maketitle

\section{Introduction}

The application of facial recognition models is controversial. Although a powerful facial recognition system can match the photo of unknowing people to help solve theft, shoplifting, and even murder cases, the misuse of facial recognition technology may pose a severe threat to individuals due to its radical erosion of privacy. For instance, a private company, Clearview.ai\cite{news}, has been reported to release a facial recognition app based on a database including more than three billion online images posted on social media. Any user can upload a picture of a person to this app to see matched public photos of that person, all without the person's consent or awareness. 

Recently, the privacy issues of facial recognition are getting more and more attention. Several tools and attacks have been proposed to help users against being identified by unauthorized facial recognition systems. In this paper, these researches are called attacks because they compromise the facial recognition system, even though they are devised to protect personal privacy.
Based on the stage where the attack occurs, these methods can be categorized into two types. One type of attack is to distort the test images to avoid being recognized at the time of inference, such as replacing faces\cite{facereplace,protect1,wu1,zhangsrame} and adding adversarial perturbations\cite{patch1,patch2,patch3,wen2021,wu2}. 
However, these methods are limited in practicality. In the real world, the test images are mostly taken by cameras and surveillance. Individuals have no access to distorting their images unless hacking the facial recognition system. Instead, another type of attack takes advantage of the clean label poisoning attack, which distorts the facial images before the training stage. Thus the facial recognition model tends to misidentify their natural pictures. For example, since the unauthorized model collects training images from social media, Alice can add imperceptible pixel-level perturbation ("cloaks") to her images before posting them online. The unauthorized model trained on these cloaked images from Alice will consistently misidentify the natural (uncloaked) picture of Alice taken by the camera. Fawkes\cite{fawkes} and Lowkey\cite{lowkey} are two popular image cloaking systems that help users to generate cloaked images to avoid being identified by unauthorized facial recognition models. 

There are very limited studies in the defense against these facial cloaking attacks. Radiya-Dixit et al.\cite{cure} has argued that these online cloaked images can not mislead a powerful model in the future. One obvious example\cite{cure} is Fawkes attack (proposed in 2020) completely fails against MagFace model\cite{magface} (released in 2021) trained on cloaked user's images. All of the attacker's unperturbed natural pictures are classified correctly during the test stage. They\cite{cure} also mentioned that a fine-tuned pre-trained ImageNet model could detect the cloaked images with high accuracy.

\subsection{PuFace}
\begin{figure}[htbp]
    \centering
    \includegraphics[width=\linewidth]{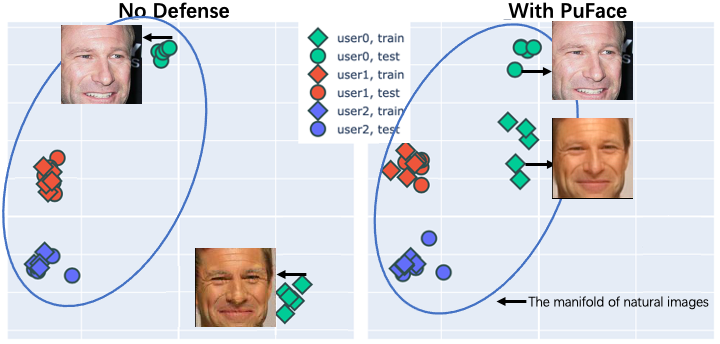}
    \caption{PCA of feature vectors extracted from facial images. The user0 uses Fawkes to cloak his train images, while user1 and user2 are regular users (both training and testing images are natural). The facial recognition model trained on all training images will misidentify the (natural) test images of user0. After PuFace purifying all the training images, the cloaked images of user0 are pulled to the manifold of natural images while the natural training images of user1 and user2 still stay there.}
    \label{f1}
\end{figure}

This paper focuses on the facial cloaking attacks happening before the training and shows that these "cloaks" are not robust enough to protect privacy since they are easy to be removed from facial images. We propose PuFace, a purification system leveraging the generalization ability of deep neural networks to push the cloaked images towards the manifold of natural images before the training process. Figure \ref{f1} presents an example of applying PuFace and also reveal the intuition behind it. The two plots show 2D principal components analysis (PCA) of feature vectors of facial images from different users. Among them, user0 (green) uses facial cloaking attacks and add some cloaks to his training images in advance, whereas user1 (red) and user2 (purple) are regular users. Their training images are uncloaked. We found the cloaked training images (green square) lie far away from the natural (uncloaked) testing images (green circle), whereas the images of user1 and user2 locates in clusters. There is no wonder that the model fails to identify the natural images of user0 as user0, referring to the left plot in Figure\ref{f1}. The right plot in Figure\ref{f1} shows the result after using PuFace. Puface successfully push these cloaked images back to the manifold where natural images lie while keeping natural training images still staying there. Therefore, the model trained on purified images can correctly identify user0 as user0.

PuFace contains a purifier that takes all the training facial images as input, regardless of whether the images is cloaked or natural, and generates purified facial images around the manifold of natural images without excessively distorting them. The purifier leverage the architecture of Residual Encoder-Decoder Network (RedNet)\cite{rednet, wen2024} to learn the mapping from the cloaked images to natural images and trained on these cloaked-natural image pairs. Moreover, we observed apparent improvement in the performance of the purifier if we train the purifier on augmented cloaked images. To push the clocked images close to the natural images, we train the purifiers to minimize two loss functions: image loss between purified images and the original natural images and feature loss between the features of purified images and original natural images extracted by an additional feature extractor. The additional feature extractor only provides the purifier with feature vectors for given images and can be discarded after training. By Minimizing both loss functions together, the purifier is encouraged to generate facial images resembling the natural images. The entire training process helps the purifier learn the representation of natural facial images and imposes a small amount of distortion to the input. Given that cloaked images lie further away from the natural images, the purifier learns to push the cloaked images onto the manifold of natural images, which can have the desirable effect of removing the cloaks.

PuFace is easy to be deployed as it does not tamper with the model's architecture, or requires a separate mechanism to detect cloaked images. We evaluate PuFace with two popular facial cloaking attacks: Fawkes\cite{fawkes} and Lowkey\cite{lowkey} with their most robust samples on six different facial recognition models. Our comprehensive experimental results present that our PuFace can effectively decrease the success rate of facial cloaking attacks without sacrificing the accuracy of the facial recognition model for regular users. For example, for 1NN model based on VGGFace2, Fawkes and Lowkey only achieves 3.29\% and 0.91\% success rate respectively with PuFace. If no defense deployed, Fawkes and Lowkey have 97.44\% and 77.83\% success rate. On average, PuFace reduce the attack success rate from 69.84\% to 7.61\%. For regular users, the classification accuracy of their natural images is still above 99\%. To our best knowledge, PuFace is the first defensive mechanism for facial cloaking attacks specifically. Therefore, we compare PuFace with several image transformation based defenses in the domain of adversarial attacks. The comparison shows that PuFace outperforms existing defenses by a considerable margin.

\subsubsection*{\textbf{Contributions}}
In Summary, this paper makes the following contributions:

\begin{itemize}

    \item We show that current facial cloaking attacks are not robust as cloaks on the facial images can be purified and removed. To users, the facial cloaking attacks provide a false sense of security. 
    
    \item We propose a common purification framework as a defense against facial cloaking attacks by pushing all the training facial images to a natural range and reduce the effect of cloaks added by these attacks.
    
    \item To achieve the defense goal and get better performance, we propose two training strategies for PuFace. 1) training the purifier on amplified cloaked images; 2) incorporating the feature loss and images loss into the training objective.
    
    \item We empirically demonstrate that PuFace effectively mitigates facial cloaking attacks without decreasing the accuracy of the facial recognition model for regular users.

\end{itemize}

\section{Background and Related Work}

\subsection{Attacks to neural networks}

Neural networks are known to be vulnerable to data poisoning attacks \cite{dp1,dp2,wu4} during the training phase and adversarial attacks\cite{adv1,adv2,wu3} happening at the inference stage. A small amount of perturbation to images that are hard to distinguish by the human can mislead the neural networks to predict wrong results.

\subsubsection{Clean label poisoning attacks\cite{dp3,dp4}}
Data poisoning attacks happen at the training stage, where an attacker injects poisoned data (correctly labeled poisoned images) into the training data set, causing a model trained on this dataset to misclassify a specific image related to this label. Unlike regular poisoning attacks, which may label a cat with a label dog, clean label poisoning attacks only add small perturbations to specific images in the training dataset and keep all the labels unchanged. Therefore, identifying clean label poisoning attacks is much more challenging.

\subsubsection{Adversarial attacks\cite{adv1,adv2,wu5}}

Adversarial attacks happen at the inference stage, where an attacker tests a standard neural network with adversarial examples (an image with small perturbation), causing the model to predict a wrong label. Unlike data poisoning attacks, adversarial attacks can attack any standard neural network without being involved in their training process. The adversarial attack literature extensively focused on developing new algorithms which can fool kinds of neural networks. 

\subsection{Facial cloaking attacks to protect privacy}

Recent research leverages and extends the above attacks broadly to protect the user from being recognized by unauthorized facial recognition models. This setting corresponds to clean-label poisoning attacks, where a user adds small cloaks to his/her images to avoid being recognized by the model trained on these cloaked images. To keep with the terminology of the data poisoning literature, we refer to the user as the attacker since the user compromises the facial recognition model to protect his/her privacy.

Facial cloaking attacks are similar to clean label poisoning attacks in terms of attack setting, which inject cloaked images into the training dataset. The algorithms to generate the cloaks adopt the idea of adversarial examples to mislead the current feature extractors. Fawkes\cite{fawkes} and Lowkey\cite{lowkey} are two popular public tools of facial cloaking attacks.

\subsubsection{Fawkes\cite{fawkes}}
The Fawkes algorithm is like generating targeted adversarial examples. Fawkes first randomly chooses a facial image from a different user as a target and tries to search the cloaks for the input close to the target user in the feature space. The targeted user serves as a landmark to help search for the cloaks that lead to significant feature space changes. During this process, the cloak itself should be bounded in a small range. To diversify the cloaked images for one user, Fawkes will apply different cloak patterns targeted at different users among the natural images of protected users.

\subsubsection{Lowkey\cite{lowkey}}
Unlike Fawkes, Lowkey improved the attacking algorithm by simultaneously attacking an ensemble of feature extractors with different backbone architectures. Lowkey leverages signed gradient descent from PGD attacks \cite{sgd} to search for an untargeted adversarial example that maximizes the distance of the natural image and the cloaked image in the feature space. In addition, the objective function combines feature vectors of the cloaked image both with and without a Gaussian blur to improve both the appearance and transferability of cloaked images.

\subsection{Related defenses}
\label{defense}

There are very sparse researches in the defense against these novel facial cloaking attacks. Given that the cloaked images generated by these attacks are highly similar to the adversarial examples, the transformation-based defense methods against adversarial examples may be used to defend facial cloaking attacks. As these defensive mechanisms are agnostic to the underlying model and attack type, we transfer these methods to defend against facial cloaking attacks. However, the experiments result in Section\ref{eva} presents they all fail to transform the cloaked images, and facial cloaking attacks still successfully mislead the facial recognition model due to the failure to capture the trait of cloaks.

\subsubsection{MagNet\cite{magnet}}

Magnet trains a reformer network based on auto-encoders to recover the detected adversarial examples to remove the adversarial noise. The auto-encoders are trained on natural images with random noise to generate the natural images.

\subsubsection{Pixel Deflection (PD)\cite{deflecting}}

This method manipulates the images without training another network. It randomly replaces some pixels in the image with randomly selected pixels from a small neighborhood. Then, the modified image is denoised by a wavelet denoiser, which smooths the image to reduce the effect of adversarial noise.

\subsubsection{Image Super-Resolution (ISR)\cite{sr}}
ISR is a two-step model-agnostic defensive framework based on super-resolution networks. It first applies wavelet denoising to suppress potential noise patterns. Then, a super-resolution network enhances pixel resolution while removing adversarial patterns for the denoised images.

\subsubsection{Defense-GAN(DGAN)\cite{defensegan}}

Defense-Gan uses a WGAN trained on natural images to learn the representation of natural images, thereby denoise adversarial examples. It is composed of two models: a generative model, which emulates the natural images, and a discriminative model that predicts if an image is real or be generated by the generative model.

\section{Methodology}

In this section, we start with the problem formulation and introduce the attack and defense settings. Next, we present the intuition behind our methodology. Then, we propose PuFace and describe the architecture, training dataset, and optimization goal. Finally, we discuss the advantages of our proposed method.

\subsection{Problem formulation}

\begin{figure}[htbp]
    \centering
    \includegraphics[width=\linewidth]{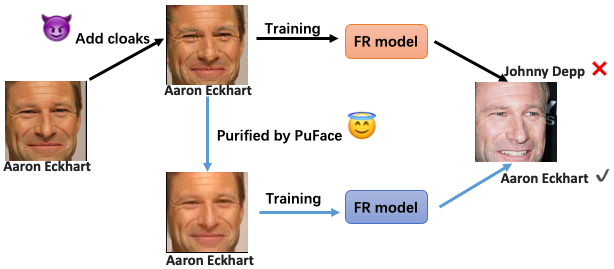}
    \caption{Attacks and defenses for facial recognition model. When the model training on Aaron Eckhart's cloaked images, it misidentifies his natural images as Johnny Depp. After PuFace purifies the training images, the model trained on purified images can correctly identify Aaron Eckhart.}
    \label{f2}
\end{figure}

First of all, we make some definitions for this problem. We consider three parties, namely facial recognition model trainer, attacker, and defender. Figure \ref{f2} presents an example of attacks and defenses for the facial recognition model.

The model trainer could be a private company like Clearview.ai, an organization, or even an individual, whose goal is to train a large-scale facial recognition model $y = \mathbf{R}(\vec{x})$, where $\vec{x}$ is a facial image while $y$ is the predicted identity.
This model aims at identifying a large number of users rather than a specific individual. With no direct access to user's personal albums, the model trainer scraps the Web for paired public facial images with names (e.g., the picture users shared on Facebook with their real name):$(\vec{x}_y,y)$ to form a labeled training dataset $\mathbf{D}_{train}$. The model trainer may either leverage the state-of-the-art feature extractors to speed up the training process or train their model from scratch.

We refer to the users who add imperceptible cloaks to their photos before posting them online as the attacker, although they intend to protect their privacy. The goal of the attacker is that their unperturbed facial images will not be identified by the facial recognition model from the model trainer, which is $a \neq \mathbf{R}(\vec{x}_a)$. We consider a strong attacker who can cloak all his facial images to be posted: $\vec{x}_a^{c} = \vec{x}_a+\vec{c}$, where $c$ is the cloak generated by Fawkes or Lowkey. So the model trainer can only collect the cloaked facial images $(\vec{x}_a^{c},a)$ of the attacker.

The defender could be the model trainer itself or a third party. In order to prevent the model from the attacks, the defender wants to purify the training facial images $\mathbf{D}_{train}$ the model trainer collected before the training process. Thus the model can be trained and work normally. The reason why our defense is effective and the detail of our defense is in the following. 

\subsection{Intuition}

The imperceptible cloaks can prevent the model from identifying the unperturbed facial images because the facial recognition model trained with these cloaked images learns the wrong features about the attacker from the cloaks. The model trainer even can not realize the attack happened as the model can correctly recognize the cloaked images of the attacker as the same person. However, the model will fail to recognize the unperturbed images of the attacker (e.g., photos taken by camera or surveillance) as the attacker because the features in the real images are completely different from what the model has learned. Intuitively, if the defender can remove these fraudulent cloaks and generate the original facial images in the training dataset, the model can then be trained and work as no attacks happen. 

Although these facial cloaking attacks happen before the training phase, and the cloaked images play the role of poisoned samples, the attacking algorithm to generate these cloaked images, such as Fawkes and Lowkey, indeed takes advantage of the idea of adversarial attacks.  Several successful defenses\cite{defensegan,magnet,sr,dcn} against adversarial attacks are motivated by the manifold  hypothesis\cite{manias,manias2}, which assumes that the training data for a single task are lying on a low dimensional manifold. The manifold hypothesis suggests that the natural images for a task are on a manifold while adding the adversarial noise to the natural images takes them off-the-manifold with high probability. Then a model that learns to push off-the-manifold adversarial images towards natural images lying on the manifold can detect and defend against adversarial attacks. 

Given that Radiya-Dixit et al. in \cite{cure} showed that the cloaked images could be distinguished from natural images with high accuracy by a binary classifier, we may assume that the cloaked image and its corresponding natural images are lying in different manifolds, despite the fact that they look almost identical. The PCA result shown in Figure \ref{f1} verifies our assumption. Inspired by these defense mechanisms in the adversarial attacks domain, we begin to consider the possibility of training a network that can generate facial images lying in the manifold of natural images from the cloaked images directly. 

\subsection{PuFace}

We propose PuFace, a simple model-agnostic facial images purification system to defend against facial cloaking attacks by pushing all the facial images in the training dataset towards a manifold where natural images lie. Our PuFace is an independent module that will not tamper with the training process and deployment of the facial recognition model. 

The simplicity of the PuFace is that it only contains a universal purifier: $\vec{\tilde{x}} = \mathbf{P}(\vec{x})$, which takes all the training images as inputs regardless of whether the image is cloaked or not. Therefore, we do not require a separate mechanism to detect cloaked images. The required characteristic for the purifier is the ability to suppress the potential cloaks and move the input to a manifold where natural images lie. In particular, the purifier achievesi the two goals: 1) for a cloaked image, the purifier aims to remove the invisible cloaks and generate a similar image that lies in a natural manifold; 2) The purifier wants to keep a natural image identical since it already lies in a natural manifold. That is to say, the ultimate goal of the purifier is to generate a similar facial image that lies in a natural manifold from natural or cloaked images.

\subsubsection{Architecture}

We adopt the architecture for our purifier from Residual Encoder-Decoder Networks(RedNet)\cite{rednet}, which have presented impressive results for image restoration tasks, such as image denoising, deblurring, inpainting, and super-resolution. 
RedNet is composed of multiple layers of symmetric convolution and deconvolution operators, which play a role in encoder and decoder, respectively. In our problem, the convolutional layers extract the input's facial features while eliminating the potential cloaks. The deconvolutional layers then recover the detail of the facial images. There are several skip connections between every two layers, from the convolutional layer to its symmetric deconvolutional layer. The skip connections pass and add feature maps extracted by convolutional layers to deconvolutional feature maps elementwise.
Therefore, necessary facial details from the convolutional feature maps can be propagated to the corresponding deconvolutional layers forwardly, which helps to recover the natural facial image. 

\subsubsection{Training dataset}
\label{training}
As the defense goal is to generate the natural facial images from cloaked images, the training set of purifier contains cloaked images $\vec{n}^c$, standing for other manifolds, and corresponding natural images $\vec{n}$, which represents the natural manifold. Fortunately, that Fawkes and Lowkey are open source tools, we were able to obtain some cloaked images to form the training dataset $\mathbb{D}_{purifier}$, which is composed of cloaked image natural image pairs. Then the purifier learns to generate natural images from cloaked images: $\vec{n} = \mathbf{P}(\vec{n}^c)$. Although the training input is only cloaked images. Our experiment shows that it does not perturb natural input excessively, which means training only on cloaked images can satisfy both goals for natural and cloaked images.
Furthermore, we find that the purifier's performance improves when the purifier is trained with stronger cloaked images, which can be obtained by amplifying the cloaks multiple times. For each cloaked image, we calculate $\vec{n}_a^{c} = \vec{n}_a+\alpha\vec{c}$, where $\alpha$ is the amplification factor. In section\ref{evanoise}, we will discuss the impact of $\alpha$ on the performance of our purifier.

\subsubsection{Optimization goal}

Learning the mapping from cloaked images to natural images needs to estimate the weights $w$ of the purifier.
Formally, given a training pair $(\vec{n}^c, \vec{n})$ in training dataset $\mathbb{D}_{p}$, where $\vec{n}^c$ is a cloaked image and $\vec{n}$ is its corresponding natural image, the optimization goal is achieved by minimizing both image loss and feature loss in Equation \ref{e1}, where $\lambda$ controls the relative importance between two loss functions during training.

\begin{equation}
\mathcal{L}(\mathbf{P}_w) = \mathcal{L}_{image}(\mathbf{P}_w) + \lambda \mathcal{L}_{feature}(\mathbf{P}_w)
\label{e1}
\end{equation}

The image loss $\mathcal{L}_{image}(\mathbf{P}_w)$ is essentially the Mean Squared Error (MSE) between the generated images by purifier and natural images, refer to Equation \ref{e2}.
MSE is a common-used loss function for image restoration. The image loss encourages the purifier to learn the mapping from cloaked images to natural images, such that the purifier can generate more natural images.

\begin{equation}
\mathcal{L}_{image}(\mathbf{P}_w) = \mathbb{E}_{\vec{n}\sim \mathbb{D}_{p}} \|\mathbf{P}(\vec{n}^c) - \vec{n} \| ^2_2
\label{e2}
\end{equation}

The feature loss $\mathcal{L}_{image}(\mathbf{P}_w)$ in Equation \ref{e3} is added to regularize the training of the purifier and preserve the key features of the facial images. During the training process, we need an additional pre-trained feature extractor $\mathbf{F}$, which can extract the feature embedding of the facial images. The feature loss measures the Mean Absolute Error (MAE) between the feature embeddings of generated images and natural images. Specifically, minimizing the loss of feature embedding encourages the generated image to retain perceptually important facial features in natural images, which are extracted by the feature extractor $\mathbf{F}$, thereby reducing the impact of cloaks in another way.
Our preliminary experiments also validate the proposal of feature loss. We observe the improvement of performance by cooperating the feature loss into the loss objective of the purifier.

\begin{equation}
\mathcal{L}_{feature}(\mathbf{P}_w) = \mathbb{E}_{\vec{n}\sim \mathbb{D}_{p}} \|\mathbf{F}(\mathbf{P}(\vec{n}^c)) - \mathbf{F}(\vec{n}) \|
\label{e3}
\end{equation}

The additional feature extractor $\mathbf{F}$ is to measure the magnitude of feature embedding for the loss objective. It will not be trained during the training process and can be discarded after training. The purifier then works independently after training.

\subsection{Advantages of PuFace}

PuFace offers several advantages. First and most important, it is agnostic to the facial recognition model and does not tamper with the training process. Second, PuFace does not require a separate mechanism to detect cloaked images that improves defense efficiency and avoids a binary detector's false-negative. Third, our defense preserves images quality while diminishing the effect of cloaks, whereas some existing defenses distort the images visibly as part of their defense. Last, PuFace can provide adequate defense without degrading the accuracy for regular users.

\section{Experimental Setup}
\subsection{Datasets}

We perform all the evaluations with two benchmark facial recognition datasets widely adopted in previous works. For simplicity, the facial images are extracted according to the official bounding box and resized to $112 \times 112$ in RGB color space, with each pixel value in the range of 0 to 1. We briefly introduce each dataset in the following.

\begin{itemize}
    \item FaceScrub \cite{facescrub} dataset contains 100K links for online facial images of 530 celebrities. We were able to download over 50K of 530 celebrities during this research, as we got access denied for some of the links. We assume FaceScrub is the training dataset that the facial recognition model provider collected from the Internet. An attacker can use either Fawkes or LowKey to cloak all of his training data. To help the facial recognition model learn about the generic data distribution for each celebrity, we discard identities that contain less than 100 images. For each user, images are divided into a training set scraped by the model provider to form $\mathbf{D}_{train}$, and a natural test set for evaluation at a 70\%-30\% split.
    
    \item CelebFaces Attributes Dataset (CelebA) \cite{celeba} contains more than 200K facial images of over 10K celebrities. We removed several celebrities whose images are also included in FaceScrub. This dataset is used to train the purifier by the defender. We assume the attacker has no access to $\mathbb{D}_{purifier}$. Thus all the images in this dataset are natural without cloaks.

\end{itemize}

Moreover, we use two non-overlapping facial image datasets in the experiments to show that the purifier can learn the manifold of natural facial images and generalize it to other facial image datasets well.

\subsection{Facial cloaking attacks}

We implemented Fawkes and Lowkey based on their open-sourced code and some public large-scale facial recognition models. As our images are cropped and aligned in advance, we skipped the alignment operation in Fawkes and LowKey.

\subsubsection{Fawkes\cite{fawkes}}

Fawkes uses VGGFace2\cite{vggface2} and WebFace\cite{webface} as the backend feature extractors. It also provides three modes, which are different tradeoff between the attack success rate (a.k.a. protection rate) and the visibility of cloaks. Our implementation for Fawkes is in high mode, which means adding more perturbation to the image and providing stronger protection.

\subsubsection{Lowkey\cite{lowkey}}

Lowkey uses ensemble of four models to generate the most robust cloaked images. The ensemble is composed of ArcFace\cite{arcface} and CosFace\cite{cosface} models. Each facial recognition model is trained with ResNet-152, and IR-152\cite{resnet} backbones respectively on the MS-Celeb-1M dataset\cite{ms}.
     
In our experiment, the attacker is a randomly chosen celebrity from FaceScrub and generates the cloaked images for all his/her facial images. Therefore, the training set collected by the model provider contains his/her cloaked images with the identity and other natural images.

\subsection{Facial recognition models}

The model provider takes advantage of a pre-trained large-scale facial feature extractor $\vec{f} = \mathbf{E}(\vec{x})$ to develop his own facial recognition system based on the training dataset collected. The pre-trained large-scale facial feature extractor takes in a facial image and predicts a vector of features. As \cite{cure} has found that both Fawkes and Lowkey achieved lower success rates on newly released models, it is more appropriate to evaluate the effectiveness of defenses on models which can be successfully attacked. Therefore, we assume the model provider uses the models used by the attacker to ensure the success of attacks. Specifically, the model provider uses VGGFace2\cite{vggface2} from Fawkes and ArcFace\cite{arcface} from Lowkey as feature extractor and trains the facial recognition model with three training strategies respectively, which previous works \cite{fawkes,lowkey,cure} also considered in their evaluations

\begin{enumerate}
    \item 1-Nearest Neighbor(1NN): 1NN is one of the oldest methods known with extremely simple idea: to identify a facial image $\vec{x}$, it finds its closest neighbor $(\vec{x'}, y')$ with smallest  Euclidean distance $\|\mathbf{E}(\vec{x})- \mathbf{E}(\vec{x'})\|_2$ in the feature space
    among the training dataset and recognizes it as $y'$.
    
    \item Linear classifier over features: a linear classifier is trained with dataset contains feature-identity pairs $(\mathbf{E}(\vec{x}),y)$. It is trained and makes predictions over the feature vector of a facial image. Therefore, the linear classifier can take advantage of powerful large-scale models and even APIs such as Azure Face from Microsoft with small calculation resources.
    
    \item Fine-tuning over extractor: the model provider adds a linear layer over the extractor $\mathbf{E}$ and fine-tune the whole model. The difference between the fine-tuning model and the linear classifier is that the extractor parameter will get updated during the fine-tuning process while the linear classifier only updates itself and uses a fixed extractor.
    
\end{enumerate}

In addition, as Fawkes and Lowkey rely on a different set of models, we are able to evaluate the transferability of two attacks across models. All in all, two feature extractors with three training strategies provide us with six different facial recognition models in the experiments.

\subsection{PuFace}

We implement PuFace using Pytorch, and all the details and related parameters will be introduced in the following. The code will be available at Github (omitted here for anonymity) to reproduce our experiments.

\subsubsection{Purifier}

The purifier is implemented based on the structure of RedNet30\cite{rednet,git}, which has 15 convolutional layers, followed by 15 deconvolutional layers and several skip connections every two layers from a convolutional layer to its symmetric deconvolutional layer.
Each convolutional and deconvolutional layer shares the same parameters: 64 feature maps, kernel size of $3\times3$, stride 2 and padding 1 to make the input and output the same size, except that the input and output feature map are set to 3. We adopt stride 2 for the first convolutional layer and the last deconvolutional layer to reduce the computational cost, which also achieves satisfactory results.
The ReLU activation function is used in every layer, including skip layers. We train the purifier by Adam optimizer with a learning rate of 0.0001 for 100 epochs.

Further, we use MegFace\cite{magface} with iResNet100 backbone as the additional feature extractor to calculation feature loss referring to Equation \ref{e3}. This feature extractor is different from the ones Fawkes and Lowkey used in their attack. $\lambda$ is set to 1. 

\subsubsection{Training dataset}

Although we have more than 200K facial images in CelebA. Generating the most robust cloaked images is very time-consuming. Due to some circumstances, we conduct all the experiments on a personal computer with a single GPU. It takes approximately 75 seconds to generate 4 Fawkes images and 30 seconds for 1 Lowkey image since Lowkey algorithm is not batched. Therefore we generate only 10K Fawkes training pairs to train the purifier. The evaluation results shows that he purifier trained on Fawkes examples generalizes well to defend against Lowkey attack as our expectation. We look forward to future research which improves the performance of Puface by using the more computational resource.

In Section \ref{training}, we mentioned that the performance of the purifier gets improved when it is trained with stronger cloaked images. In our experiments, we set $\alpha = 5$ for the best performance, which means the cloaks are strengthened to 5 times and added to the natural images. Moreover, in Section\ref{evanoise}, we will present a comprehensive evaluation in terms of the amplification factor $\alpha$.

\subsection{Existing Defenses}

We compare our PuFace with the following existing image transformation-based defense as introduced in Section \ref{defense}, which represent the state-of-the-art defense in the domain of adversarial attacks in the literature. We share the same idea with these defense methods: to transform the adversarial examples, in our scenario, the cloaked images into more natural images.

\subsubsection{MagNet\cite{magnet}}

We use the open-source code to implement MagNet and apply its reformer for reconstructing the whole training dataset. The reformer is trained on noisy facial images from CelebA with the objective to generate clean facial images.  

\subsubsection{Pixel Deflection (PD)\cite{deflecting}}

We adopt its open-source code to implement this defense method. We also use its default parameters: 200 pixels to be deflected with a window size of 10. Then the output is denoised by the wavelet denoiser with $\sigma = 0.04$.

\subsubsection{Image Super-Resolution (ISR)\cite{sr}}

ISR transforms the images by two steps. First, we apply the wavelet denoiser to the input with $\sigma = 0.02$. Next, we use the official model of Enhanced Deep Super-Resolution (EDSR) \cite{edsr} network trained on the DIVerse 2K resolution image (DIV2K) dataset \cite{srdata} to improve the quality of the denoised images with scale 2.

\subsubsection{Defense-GAN(DGAN)\cite{defensegan}}
We implement DGAN using its open-source code. The GAN is trained on the natural facial images from CelebA in an unsupervised manner. Then we use DGAN to transform the whole training dataset.

\section{Evaluations}

In this section, we start with the evaluation metrics. Next, we demonstrate the defense performance of PuFace and compare it with existing defenses. Next, we present the reconstructed images of different defensive methods. Last, we show how defense performance varies with the amplification factor $\alpha$ in the training dataset of the purifier.

\subsection{Evaluation Metrics}

\begin{table*}[hbtp]
\caption{The normal accuracy (left) and attack success rate (right) when different defense methods are deployed. Each row shows the result of an independent facial recognition system attacked by Fawkes or Lowkey. The lower success rate indicates a better dofense. }
\label{t1}
\begin{tabular}{|c|c|c|c|c|c|c|c|c|}
\hline
 & Model & Attacks & No Defense & MagNet & PD & ISR & DGAN & \textbf{PuFace} \\ \hline
 &  & Fawkes & 99.52\% / 97.44\% & 92.28\% / 78.75\% & 99.42\% / 85.93\% & 99.26\% / 96.81\% & 99.39\% / 91.15\% & \textbf{99.71\% / 3.29\%} \\
 & \multirow{-2}{*}{1NN} & \cellcolor[HTML]{DAE8FC}Lowkey & \cellcolor[HTML]{DAE8FC}99.70\% / 77.83\% & \cellcolor[HTML]{DAE8FC}95.58\% / 57.26\% & \cellcolor[HTML]{DAE8FC}99.85\% / 56.25\% & \cellcolor[HTML]{DAE8FC}99.56\% / 77.15\% & \cellcolor[HTML]{DAE8FC}99.55\% / 67.16\% & \cellcolor[HTML]{DAE8FC}\textbf{99.55\% / 0.91\%} \\ \cline{2-9} 
 &  & Fawkes & 99.26\% / 99.88\% & 97.83\% / 83.21\% & 99.40\% / 98.30\% & 99.23\% / 99.72\% & 99.69\% / 86.33\% & \textbf{99.54\% / 23.69\%} \\
 & \multirow{-2}{*}{Linear} & \cellcolor[HTML]{DAE8FC}Lowkey & \cellcolor[HTML]{DAE8FC}99.58\% / 71.65\% & \cellcolor[HTML]{DAE8FC}94.26\% / 63.95\% & \cellcolor[HTML]{DAE8FC}99.11\% / 57.49\% & \cellcolor[HTML]{DAE8FC}99.19\% / 75.63\% & \cellcolor[HTML]{DAE8FC}99.70\% / 63.44\% & \cellcolor[HTML]{DAE8FC}\textbf{99.55\% / 1.20\%} \\ \cline{2-9} 
 &  & Fawkes & 97.65\% / 91.33\% & 91.43\% / 61.22\% & 98.13\% / 63.33\% & 98.13\% / 76.67\% & 97.98\% / 56.67\% & \textbf{98.44\% / 10.02\%} \\
\multirow{-6}{*}{\begin{tabular}[c]{@{}c@{}}
 \rotatebox{90}{VGGFace2}\end{tabular}} & \multirow{-2}{*}{FT} & \cellcolor[HTML]{DAE8FC}Lowkey & \cellcolor[HTML]{DAE8FC}98.27\% / 78.41\% & \cellcolor[HTML]{DAE8FC}96.54\% / 23.33\% & \cellcolor[HTML]{DAE8FC}98.76\% / 16.67\% & \cellcolor[HTML]{DAE8FC}98.29\% / 10.03\% & \cellcolor[HTML]{DAE8FC}99.07\% / 20.09\% & \cellcolor[HTML]{DAE8FC}\textbf{98.60\% / 0.34\%} \\ \hline
 &  & Fawkes & 99.11\% / 19.48\% & 96.75\% / 11.68\% & 99.69\% / 12.60\% & 99.68\% / 19.03\% & 99.55\% / 10.57\% & \textbf{99.55\% / 0.89\%} \\
 & \multirow{-2}{*}{1NN} & \cellcolor[HTML]{DAE8FC}Lowkey & \cellcolor[HTML]{DAE8FC}99.41\% / 91.21\% & \cellcolor[HTML]{DAE8FC}98.26\% / 74.09\% & \cellcolor[HTML]{DAE8FC}99.70\% / 77.90\% & \cellcolor[HTML]{DAE8FC}99.55\% / 93.06\% & \cellcolor[HTML]{DAE8FC}99.84\% / 91.27\% & \cellcolor[HTML]{DAE8FC}\textbf{99.56\% / 21.69\%} \\ \cline{2-9} 
 &  & Fawkes & 99.85\% / 18.33\% & 97.56\% / 14.33\% & 99.85\% / 7.08\% & 99.41\% / 17.93\% & 99.55\% / 7.34\% & \textbf{99.21\% / 1.09\%} \\
 & \multirow{-2}{*}{Linear} & \cellcolor[HTML]{DAE8FC}Lowkey & \cellcolor[HTML]{DAE8FC}99.41\% / 99.40\% & \cellcolor[HTML]{DAE8FC}98.22\% / 95.86\% & \cellcolor[HTML]{DAE8FC}99.50\% / 94.53\% & \cellcolor[HTML]{DAE8FC}99.68\% / 99.43\% & \cellcolor[HTML]{DAE8FC}99.57\% / 93.96\% & \cellcolor[HTML]{DAE8FC}\textbf{99.83\% / 24.48\%} \\ \cline{2-9} 
 &  & Fawkes & 99.28\% / 21.43\% & 97.19\% / 33.33\% & 98.29\% / 43.33\% & 97.67\% / 23.33\% & 97.67\% / 47.00\% & \textbf{97.36\% / 3.33\%} \\
\multirow{-6}{*}{\begin{tabular}[c]{@{}c@{}}
\rotatebox{90}{ArcFace}\end{tabular}} & \multirow{-2}{*}{FT} & \cellcolor[HTML]{DAE8FC}Lowkey & \cellcolor[HTML]{DAE8FC}98.57\% / 72.09\% & \cellcolor[HTML]{DAE8FC}98.22\% / 76.67\% & \cellcolor[HTML]{DAE8FC}98.44\% / 26.67\% & \cellcolor[HTML]{DAE8FC}98.91\$ / 53.33\% & \cellcolor[HTML]{DAE8FC}98.76\% / 20.23\% & \cellcolor[HTML]{DAE8FC}\textbf{98.91\% / 0.76\%} \\ \hline
\end{tabular}

\end{table*}
\subsubsection{Defending against cloaking attacks} 

First and most important, we evaluate the effectiveness of PuFace using two metrics:

\begin{itemize}
    \item \textit{Attack success rate} (a.k.a, protection rate) of Fawkes and Lowkey is the error rate (top-1 accuracy) of the facial recognition model for natural test images of the attacker. We repeat each attack algorithm 20 times for one facial recognition model with a different identity (randomly chosen from FaceScrub) in the position of the attacker. Then we average the protection rates over 20 attackers. The lower the protection rate, the better performance of the defensive mechanism.
    \item \textit{Normal accuracy} is the accuracy of the facial recognition model for all the natural test images besides the attacker. As our PuFace purifies all the training images, including uncloaked natural images, it may degrade the performance of the facial recognition model. The higher the accuracy, the smaller the influence of the defense mechanism on the face recognition model.
\end{itemize}

\subsubsection{Evaluating purified facial images}

Purified facial images are evaluated from the following perspectives. All the following metrics are calculated between two images. In our scenario, the first input is the test images, including purified images, cloaked images, and generated images by other defensive methods, and the second input is the fixed original natural images. For simplicity, we only indicate the first input and omit the second natural image in the following tables.

\begin{itemize}

    \item \textit{Image distortion} is measured by Mean Squared Error (MSE) between the test images and the original natural images. We expect the defense introduction minor image distortion to natural images while purifying the cloaked images.
    
    \item \textit{Feature loss} is calculated using Mean Absolute Error (MAE) between the feature embeddings of test images and original natural images. Compared with the feature loss of cloaked images without defense, it indicates if the defense pushes the images towards the manifold of natural images or not.
\end{itemize}

\subsection{Experimental Results}
\label{eva}

In this section, we first present the experimental results of defense performance and compare PuFace to other defense methods in Table \ref{t1}.

Table\ref{t1} presents the normal accuracy and attack success rate for different facial recognition models attacked by Fawkes or Lowkey. Each row shows the result of an independent facial recognition system attacked by Fawkes or Lowkey. For example, the first row means the model provider trains a 1-nearest neighbor model based on VGGFace2 feature extractor and is attacked by Fawkes. The last row indicates a fine-tuning model based on ArcFace feature extractor attacked by Lowkey. In each cell, we show the model's normal accuracy, followed by the attack's success rate. For each facial recognition model, we randomly chose one celebrity from FaceScrub as the attacker and added cloaks to all the images of the attacker in the training dataset. Next, we apply each defense method and generate a training dataset with defense, respectively. For each dataset, we train a facial recognition model and calculate the normal accuracy of the model and the success rate of the attack. We repeat the attacks and training process 20 times with different identities in the position of attackers and average the normal accuracy and attack success rate over 20 experiments. Finally, the results are displayed in one cell. A strong defense method is expected to lower the attack success rate without reducing the normal accuracy.

First of all, it is evident that our PuFace dramatically reduce the attack success rate of Fawkes and Lowkey for all types of facial recognition models while the normal accuracy rate remains almost at the same level. To get a straightforward result,  we average all the results across models and attacks. The overall attack success rate drops from 69.83\% to 7.61\% with our defense. When considering the model and the attack, the defense performance shows some differences. For example, for the linear model based on VGGFace2, PuFace reduce the success rate of Fawkes from 99.88\% to 23.69\%, while reducing the success rate of Lowkey from 71.65\% to 1.2\%. One of the reasons is that Fawkes has higher success rates for models based on VGGFace2 than Lowkey, given that VGGFace2 is one of the models used by Fawkes. Consequently, defending against Fawkes for VGGFace2 is more difficult. Similarly, our defense against Lowkey for ArcFace also performs slightly worse than Fawkes.

Second, PuFace significantly outperforms other transformation-based defenses and is the most effective defense against Fawkes and Lowkey in our comprehensive experiments. Although these transformation-based defenses effectively reconstruct adversarial examples, they cannot remove the heavy cloaks generated based on an ensemble of models. This result is not surprising as they are not explicitly designed for facial cloaking attacks. In most cases, these four defenses can slightly reduce the attack success rate. However, we find a strange result. For a fine-tuning model on ArcFace, the Fawkes attack itself only has a 21.43\% success rate. These defenses increase the success rate and even double the success rate after reconstructing the facial images. In the meantime, our PuFace still works and reduces the success rate to 3.33\%. Besides, we also observed that MagNet has a noticeable impact on the normal accuracy, while PD, ISR DGAN, and our PuFace are able to retain the normal accuracy of the facial recognition model. This may be because the reformer in MagNet is explicitly designed for adversarial examples as MagNet has a detector to filter the normal examples out and reforms adversarial examples only. In our experiment, we apply the reformer of MagNet to all the training images, including natural images according to the defense setting.
\begin{figure*}[htbp]

\begin{subfigure}{0.5\textwidth}
\includegraphics[width=\linewidth]{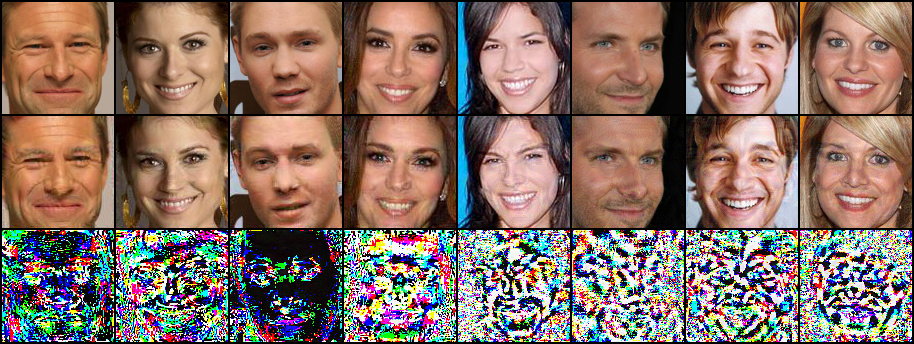}
\caption{The natural images(first row), cloaked image(second row) and the cloaks added by attackers(last row). The left four images are attacked by Fawkes while the right four images are attacked by Lowkey.}
\label{fig:a}
\end{subfigure}\hspace*{\fill}
\hfill
\begin{subfigure}{0.5\textwidth}
\includegraphics[width=\linewidth]{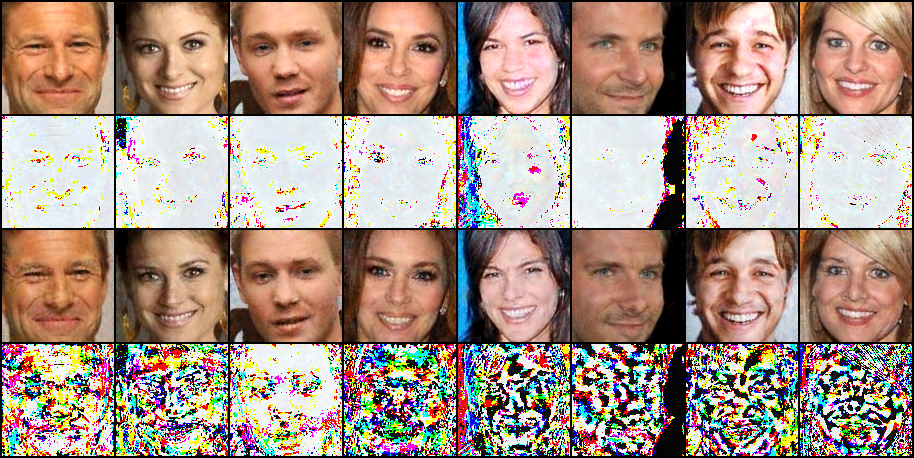}
\caption{MagNet}
\label{fig:b}
\end{subfigure}

\medskip
\begin{subfigure}{0.5\textwidth}
\includegraphics[width=\linewidth]{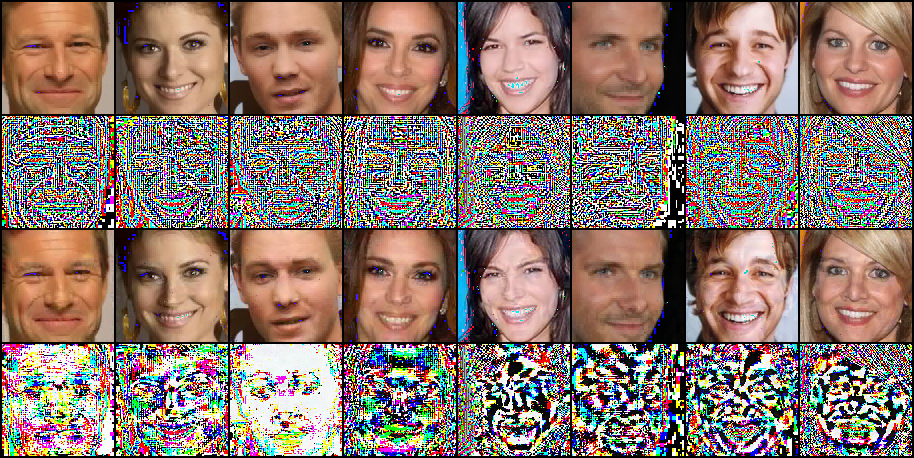}
\caption{PD}
\label{fig:c}
\end{subfigure}\hspace*{\fill}
\begin{subfigure}{0.5\textwidth}
\includegraphics[width=\linewidth]{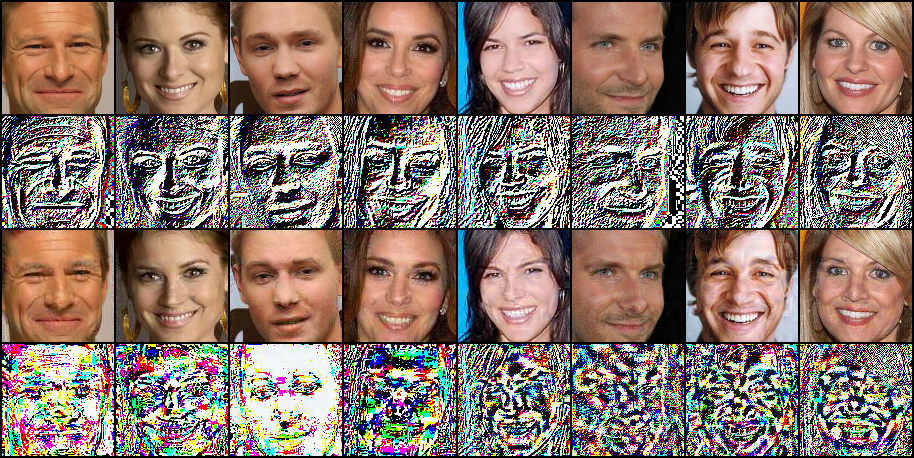}
\caption{ISR}
\label{fig:d}
\end{subfigure}

\medskip
\begin{subfigure}{0.5\textwidth}
\includegraphics[width=\linewidth]{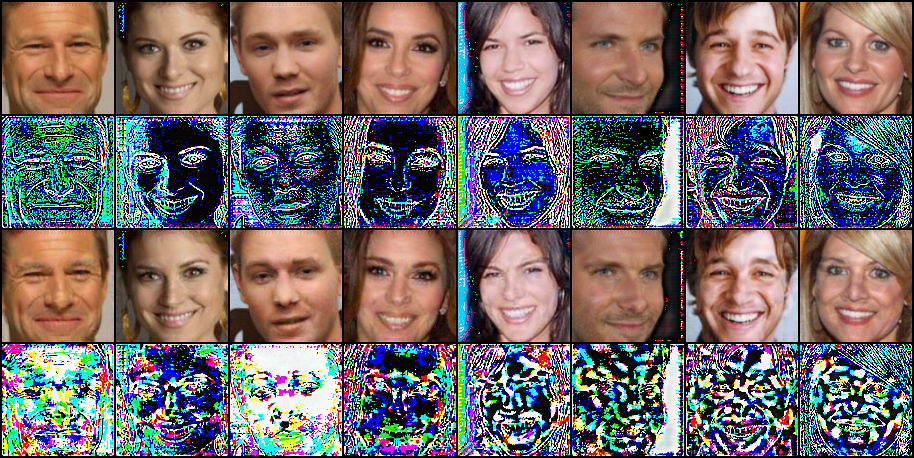}
\caption{DGAN}
\label{fig:e}
\end{subfigure}\hspace*{\fill}
\begin{subfigure}{0.5\textwidth}
\includegraphics[width=\linewidth]{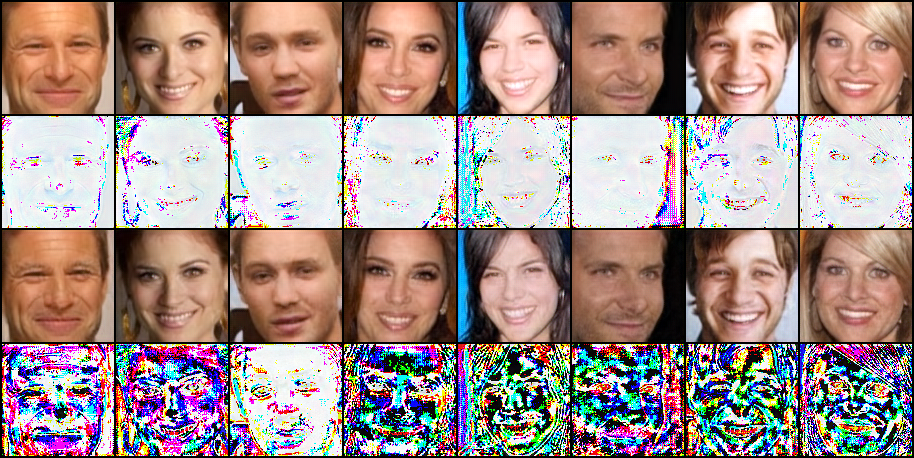}
\caption{PuFace}
\label{fig:f}
\end{subfigure}

\caption{The reconstructed results of different defenses. Figure \ref{fig:a} shows the natural images, cloaked images and the cloaks between them. Following are 5 visual comparison of different defenses. In each figure, the first and third row present the reconstructed images of natural and cloaked images respectively, and the second and last row show the difference between them and the natural images.}
\label{feva}
\end{figure*}

Next, the results in Table \ref{t1} also give us some insight into the transferability of Fawkey and Lowkey. Both attacks have a lower success rate across models. More specifically, Fawkes achieves more than 90\% success rate when attacks the VGGFace2 model which the attacks is designed for, while the same attacks have a success rate of around 20\% on the new ArcFace model the attacker has never seen before. Lowkey seems to transfer better than Fawkes. Lowkey achieves more than 70\% success rate on the new model VGGFace2, which is not included in the ensemble of attacking models. Moreover, our PuFace usually has better defensive performance against cross-model attacks.

Last but not least, as we mentioned, we train our Puface with all Fawkes samples. This is because Fawkes samples can be generated in batches, whereas Lowkey runs very slowly on our device. Despite that PuFace has never seen Lowkey samples before, it is still very effective in defending against Lowkey. This indicates Puface can map the input to the natural images, regardless of whether the input is Fawkes cloaked, Lowkey cloaked, or natural. Besides, the result also encourages us to explore the potential capability of PuFace, such as defending against adversarial examples, which might be future work.

\subsection{Reconstruction Performance}

After evaluating the defense performance, in this section, we show some randomly chosen samples of purified images for visual comparison of different reconstructed facial images by defense methods in Figure \ref{feva}. Moreover, we evaluate the reconstructed facial images in terms of image distortion and feature loss in Table\ref{t2}.

\begin{table}[htbp]
\centering
\caption{The image distortion and feature loss of different defense methods.}
\label{t2}
\begin{tabular}{|c|cc|cc|}
\hline
 & \multicolumn{2}{c|}{Image distortion} & \multicolumn{2}{c|}{Feature loss} \\ \cline{2-5} 
\multirow{-2}{*}{} & \multicolumn{1}{c|}{Natural} & Cloaked & \multicolumn{1}{c|}{Natural} & Cloaked \\ \hline
No defense & \multicolumn{1}{c|}{0} & 0.0015 & \multicolumn{1}{c|}{0} & 0.0556 \\ \hline
\rowcolor[HTML]{DAE8FC} 
MagNet & \multicolumn{1}{c|}{\cellcolor[HTML]{DAE8FC}0.0053} & 0.0024 & \multicolumn{1}{c|}{\cellcolor[HTML]{DAE8FC}0.0458} & 0.0474 \\ \hline
PD & \multicolumn{1}{c|}{0.0033} & 0.0049 & \multicolumn{1}{c|}{0.0076} & 0.0516 \\ \hline
\rowcolor[HTML]{DAE8FC} 
ISR & \multicolumn{1}{c|}{\cellcolor[HTML]{DAE8FC}0.0005} & 0.0022 & \multicolumn{1}{c|}{\cellcolor[HTML]{DAE8FC}0.0015} & 0.0552 \\ \hline
DGAN & \multicolumn{1}{c|}{0.0019} & 0.0032 & \multicolumn{1}{c|}{0.0041} & 0.0524 \\ \hline
\rowcolor[HTML]{DAE8FC} 
\textbf{PuFace} & \multicolumn{1}{c|}{\cellcolor[HTML]{DAE8FC}\textbf{0.0056}} & \textbf{0.0021} & \multicolumn{1}{c|}{\cellcolor[HTML]{DAE8FC}\textbf{0.0128}} & \textbf{0.0392} \\ \hline
\end{tabular}
\end{table}

In Figure \ref{feva}, the first Figure \ref{fig:a} shows the natural images, cloaked images, and the cloaks between them. Each defense method will process natural images and cloaked images and generate a new version of them. The eight celebrities are randomly chosen from FaceScrub, and the left four of them are attacked by Fawkes while Lowkey attacks the right four images. The third row shows the image distortion, the cloaks from Fawkes and Lowkey differ from each other a lot. Personally, both cloaks from Fawkes and Lowkey are somehow visible for us. We could see noticeable lines which look like weird wrinkles on the faces, compared with the natural images. However, these lines would not affect the recognition of the person in the image for humans. We still think these cloaked images look a little bit strange.

Following are 5 sets of reconstructed images from different defenses. In each figure, the first and third row presents the reconstructed images of natural and cloaked images, respectively, and the second and last row shows the difference between them and corresponding natural images. We analyze the reconstructed images of the defense one by one below.

\begin{itemize}
    \item For MagNet in Figure \ref{fig:b}, some visible cloaks still can be seen on the reconstructed cloaked images. The result shows that MagNet trained on natural images with noise is not sensitive to the cloaks and therefore fails to remove them from images. These cloaks are carefully crafted and embedded in the images, blending into the face better than noise.
    
    \item As for Pixel Deflection in Figure \ref{fig:c}, we could see several remarkable colorful dots on both reconstructed natural and cloaked images. Still, the flipping within the local area fails to break the patterns of cloaks. The difference between reconstructed cloaked images and natural images is very similar to the difference in Figure \ref{fig:a}.
    
    \item Image Super-Resolution indeed improves image quality by sharpening facial contour presented in the second row of Figure \ref{fig:d}. It generates more vivid reconstructed natural images. However, it still fails to capture the cloaks, which usually appear on the face instead of the edge. Super-resolution does not help to remove the impact of cloaks.
    
    \item The reconstructed results of Defense-GAN in Figure \ref{fig:e} differs from previous results. For natural images, DGAN adds more details to the eyes, nose, and mouth. However, we also notice some inconsistent pixels around the edges of images. For cloaked images, the generative power of GAN is not capable enough of suppressing the effect of cloaks.  The visual result responds to the high attack success rate in Table \ref{t1}

    \item The last Figure \ref{fig:f} show the purified images of our PuFace. First of all, the purified images look brighter and smoother. It seems that the purifier pays more attention to the skin's texture so that the cloaks can be removed. This may be because the stronger cloaked images in the training dataset supervise the purifier to learn the mapping from the cloaked images to natural images. Moreover, the feature loss and images loss helps the purifier to minimize the cloaks. Besides, PuFace introduces minor distortion on natural images, referring to the second row in Figure \ref{fig:f}.
\end{itemize}

Table \ref{t2} shows the quantitative image distortion and feature loss for the reconstructed images from various defense methods. Our PaFace successes in reducing the feature loss between the purified cloaked images and natural images. Therefore it is effective in defending against facial cloaking attacks. We also notice that PaFace introduces the most distortion to the purified natural images, compared with other defensive methods. However, the result in Table\ref{t1} reflects that these distortions on purified natural images have no impact on the normal accuracy of the model. We conclude that these distortions are acceptable and negligible.

\subsection{Train PuFace with stronger cloaked images}
\label{evanoise}

\begin{figure}[htbp]
\centering
\begin{subfigure}{0.49\textwidth}
\includegraphics[width=\linewidth]{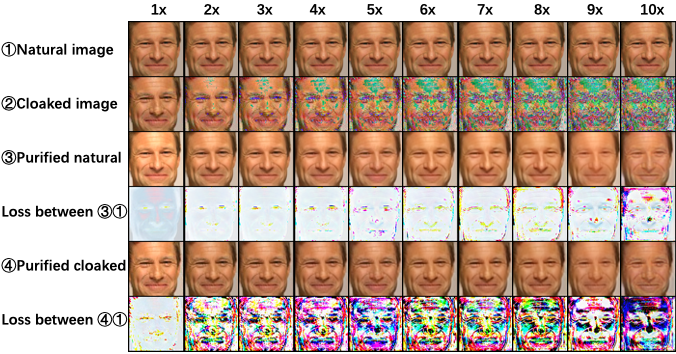}
\caption{Visual comparison}
\label{fn1}
\end{subfigure}\hspace*{\fill}

\medskip
\begin{subfigure}{0.24\textwidth}
\includegraphics[width=\linewidth]{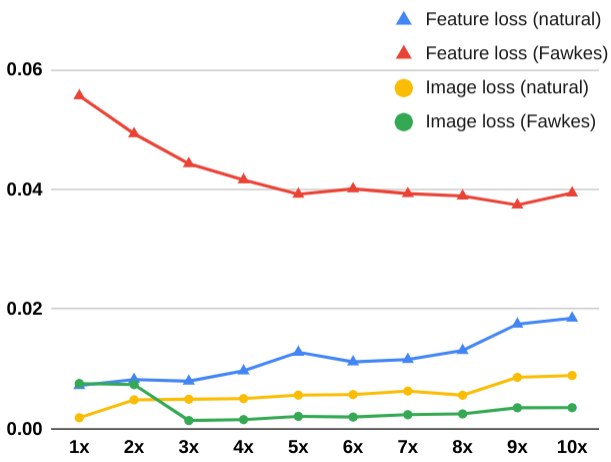}
\caption{Feature loss and image loss}
\label{fn2}
\end{subfigure}\hspace*{\fill}
\begin{subfigure}{0.24\textwidth}
\includegraphics[width=\linewidth]{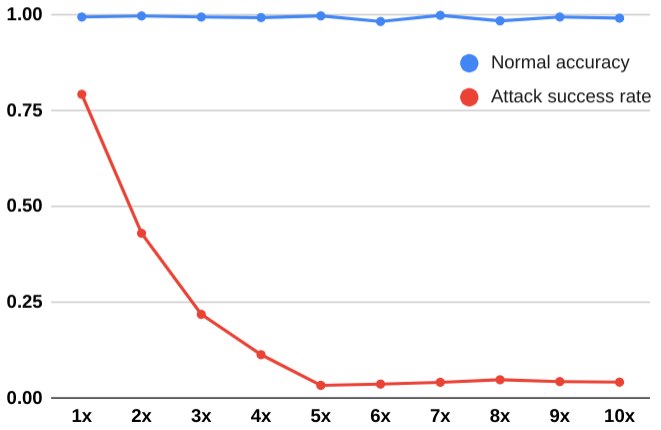}
\caption{Defence performance}
\label{fn3}
\end{subfigure}

\caption{ We train PuFace on 1-10X cloaks to show how the defense performance and purified images vary.}
\label{fn}
\end{figure}

In Section \ref{training}, we mentioned that the performance of the purifier gets improved when it is trained on stronger cloaked images. As we have already generated the most robust cloaked image within the capabilities of Fawkes and Lowkey, we further strengthen the cloaked images in the training dataset of purifier by amplifying the cloaks multiple times and applying them to the images directly.
We amplify the cloaks from 1x to 10x, increase by 1x each time, and train PuFace on these datasets, respectively. Figure \ref{fn} presents the comprehensive results for the performance of purifiers trained on datasets with different values of the amplification factor $\alpha$. In this section, for simplicity, we only evaluate the 1NN model based on VGGFace2, and it is attacked by Fawkes, as we want to pay more attention to the impact of the amplification factor $\alpha$ in the training dataset of the purifier.

Figure \ref{fn1} demonstrates how the defense performance varies from different values of amplification factor $\alpha$. The first row is the natural images, which are the same since the natural image is the ground truth image the purifier wants to generate. The second row is the cloaked images with different $\alpha$, from which the purifier learns to generate the natural images. Noted that we use the same facial image here to present the training pairs and testing result for simplicity, in our experiment,  we use facial images from CelebA to train the purifier and test with facial images from FaceScrub. These two datasets do not overlap at all. With the increase of $\alpha$, we could notice that the cloak mainly focuses on the face's eyes, nose, mouth, and forehead. We began to think that using heavily cloaked images may help the purifier pay more attention to these areas where cloaks usually appear and learn to recover the natural images. The results verified our guess. In Figure \ref{fn3}, we show how the normal accuracy and attack success rate varies with $\alpha$. When training the purifier with the original cloaked images, Fawkes still achieves an 83.21\% success rate. With the increase of $\alpha$, the success rate drops rapidly, staying at around 4\% ($\alpha \geq 5)$, with normal accuracy almost unchanged. The trend of success rate is identical to the feature loss between the purified cloaked image and the natural images (red line) in Figure \ref{fn2}. That is to say, learning with heavily cloaked images, the purifier can push the cloaked images closer to the natural images in the feature space, which meets our defense goal. On the other hand, the purifier also introduces more distortion when purifying natural images. Both the image loss and feature loss between the purified natural images and original natural images increase slightly with the increase of $\alpha$. In conclusion, $\alpha$ actually is a tradeoff between the defense performance and the quality of purified images. For instance, in Figure \ref{fn1}, the purified facial images become visibly pale and blurred when $\alpha > 5$, despite that the success rate is low. Considering the results in Figure \ref{fn}, we set $\alpha = 5$ for PuFace to balance the defense performance and the quality of purified images.

\section{Conclusions}

This paper proposed PuFace, a simple add-on defense strategy to defend against facial cloaking attacks by purifying the training dataset and removing potential cloaks. The core idea is to train a purifier that pushes the input towards the manifold of natural images. To achieve this goal, we further proposed to train the purifier on amplified cloaked images and minimize both image loss and feature loss together. We empirically show that PuFace can consistently provide adequate defense, reducing the attack success rate from 69.84\% to 7.61\% for kinds of models without sacrificing the normal accuracy. Moreover, PuFace is compatible with any model as it is an easy pre-processing step for training datasets prior to classification.

Our work also shows that these facial cloaking attacks deliver a false sense of security for those who want to protect their privacy relying on these tools. Therefore, we believe that it is legislation rather than techniques that is needed to protect our privacy from unauthorized facial recognition systems.


\bibliographystyle{ACM-Reference-Format}
\bibliography{sample}


\end{document}